\documentclass[fleqn,10pt]{wlscirep}
\usepackage[utf8]{inputenc}
\usepackage[T1]{fontenc}
\usepackage{amsmath}
\usepackage{nccmath}
\usepackage{bbm}
\usepackage{bm}
\usepackage{xcolor}

\usepackage[T1]{fontenc}
\usepackage{caption}
\usepackage{caption,booktabs}

\title{Uncovering spatial tissue domains and cell types in spatial omics through cross-scale profiling of cellular and genomic interactions}

\author[1,2]{Rui Yan}
\author[2]{Xiaohan Xing}
\author[3]{Xun Wang}
\author[2]{Zixia Zhou}
\author[2]{Md Tauhidul Islam}
\author[1,2,*]{Lei Xing}
\affil[1]{Institute for Computational and Mathematical Engineering, Stanford University, Stanford, CA 94305, USA}
\affil[2]{Department of Radiation Oncology, Stanford University, Stanford, CA 94305, USA}
\affil[3]{Gladstone Institute of Cardiovascular Disease, San Francisco, CA 94158, USA}
\affil[*]{Corresponding author: lei@stanford.edu}

\begin{abstract}
Cellular identity and function are linked to both their intrinsic genomic makeup and extrinsic spatial context within the tissue microenvironment.
Spatial transcriptomics (ST) offers an unprecedented opportunity to study this, providing \textit{in situ} gene expression profiles at single-cell resolution and illuminating the spatial and functional organization of cells within tissues. 
However, a significant hurdle remains: ST data is inherently noisy, large, and structurally complex. This complexity makes it intractable for existing computational methods to effectively capture the interplay between spatial interactions and intrinsic genomic relationships, thus limiting our ability to discern critical biological patterns.
Here, we present CellScape, a deep learning framework designed to overcome these limitations for high-performance ST data analysis and pattern discovery. 
CellScape jointly models cellular interactions in tissue space and genomic relationships among cells, producing comprehensive representations that seamlessly integrate spatial signals with underlying gene regulatory mechanisms. 
This technique uncovers biologically informative patterns that improve spatial domain segmentation and supports comprehensive spatial cellular analyses across diverse transcriptomics datasets, offering an accurate and versatile framework for deep analysis and interpretation of ST data.  
\end{abstract}

\begin{document}
\flushbottom
\maketitle

\vspace{-2em}
\section*{Introduction}
Multicellular tissues are complex systems in which individual cells interact, coordinate, and specialize to perform higher-order biological functions. Each cell acts as a dynamic unit, with its behaviors governed by a myriad of intrinsic molecular interactions and modulated by extrinsic signals from the local cellular environment. The coordinated interplay among diverse cell populations contributes to the emergence of tissue-scale architecture and function~\cite{rao2021exploring}. 
Recent advances in spatial transcriptomics (ST) have enabled the simultaneous profiling of gene expression and spatial localization of individual cells within the tissue~\cite{codeluppi2018spatial,stickels2021highly,chen2015spatially,moses2022museum}, opening new avenues to investigate cellular composition and activity in situ~\cite{marx2021method}. However, extracting meaningful biological insights from these datasets remains challenging due to the inherent multi-scale complexity of tissues, arising from vast interacting components across genomic, cellular, and spatial dimensions.

To effectively analyze ST data, it is essential to consider both the intrinsic genomic regulation governing cellular states and the spatial organization that enables intercellular interactions~\cite{chidester2023spicemix,dong2025simvi}.
Traditional approaches such as Louvain clustering, were originally designed for single-cell RNA sequencing (scRNA-seq). These methods define cell identity based solely on transcriptomic similarity, ignoring spatial information and limiting their ability to capture biologically meaningful spatial patterns in tissues~\cite{moffitt2018molecular,eng2019transcriptome}.
Recently, a range of deep learning methods~\cite{long2023spatially,liang2024prost,hu2021spagcn,dong2022deciphering,xu2024unsupervised,varrone2024cellcharter,ren2022identifying} based on graph neural networks (GNNs) and autoencoders, has emerged to integrate gene expression with spatial information.
For instance, SpaGCN~\cite{hu2021spagcn}  uses graph convolutional networks to jointly model spatial location and gene expression for unsupervised clustering, and GraphST~\cite{long2023spatially} employs contrastive learning to learn representations by considering both spatial and transcriptional similarity of the cells.
However, these approaches tend to overemphasize spatial adjacency and obscure biologically meaningful transcriptomic differences, often leading to compromised results~\cite{chidester2023spicemix}. 
Furthermore, current methods often fail to capture higher-order intracellular features, such as gene–gene interactions. Incorporating these co-expression patterns is crucial for revealing cellular behaviors and identities~\cite{van2018gene}, thereby enhancing ST analysis.
In addition, many existing frameworks lack built-in mechanisms for correcting batch effects, leading to suboptimal performance in multi-sample spatial omics analyses.

\begin{figure}[t]
\centering 
\includegraphics[width=\linewidth]{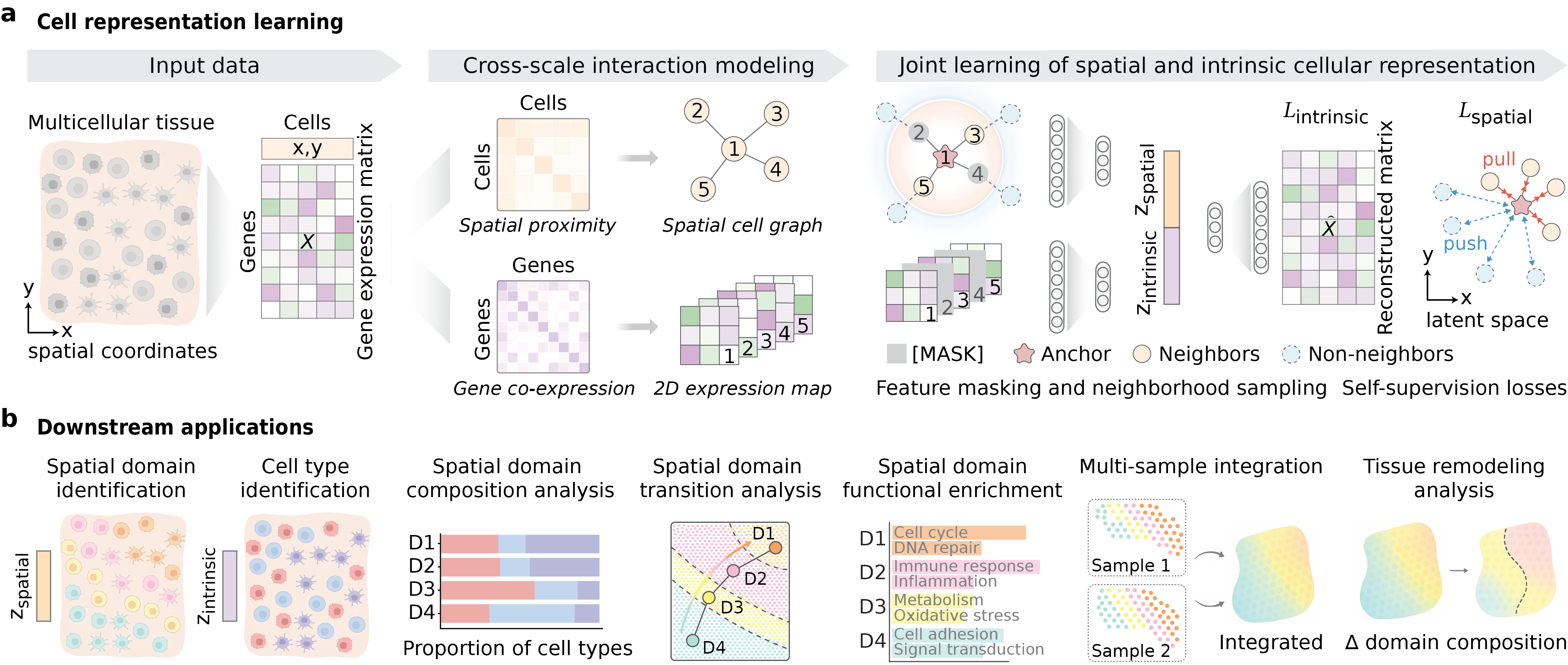}
\caption{\textbf{CellScape characterizes cells via joint modeling of spatial and genomic interactions.}
\textbf{a}, Workflow of CellScape. Given a gene expression matrix and spatial coordinates for each cell in the tissue, CellScape constructs a cell graph based on spatial proximity and a 2D gene expression map reflecting gene co-expression patterns. Its dual-branch architecture employs two encoders: one captures spatial interactions and produces spatial embeddings $Z_{\text{spatial}}$, and the other captures intrinsic gene expression structure that produces intrinsic embeddings $Z_{\text{intrinsic}}$. Each representation provides a distinct view of the cellular landscape and can be leveraged for task-specific analyses.
\textbf{b}, CellScape enables various downstream tasks for spatial omics data analysis.
} 
\label{Fig:framework}
\vspace{-1em}
\end{figure}

Here, we present CellScape, a dual-branch deep learning framework that synergistically reveals the spatial and functional organization of individual cells within tissues. 
Two encoders are synergistically integrated in CellScape: one prioritizing spatial interactions in tissue space, and the other emphasizing genomic relationships within each cell (Fig.~\ref{Fig:framework}a). CellScape jointly models spatial and genomic interactions through self-supervised learning (see Methods). The resulting representations offer complementary views of the transcriptomics data and can be used independently for task-specific analyses.
We demonstrate that CellScape accurately identifies spatial domains, enabling a range of downstream ST analyses and applications (Fig.~\ref{Fig:framework}b). This includes profiling domain-specific cell type composition, examining spatial domain transitions, and characterizing spatially organized cellular functions.
The framework also corrects batch effects to enable robust multi-sample integration for comparative tissue remodeling analysis across conditions.
We further showcase CellScape’s versatility by using datasets with varying resolutions generated by multiple ST technologies. Remarkably, CellScape consistently outperforms existing methods in spatial domain segmentation and reveals biologically meaningful patterns with high fidelity.

\section*{Results}

\subsection*{CellScape enables spatially resolved cell-type-specific gene expression analysis}
\begin{figure}[!t]
\centering
\includegraphics[width=0.95\linewidth]{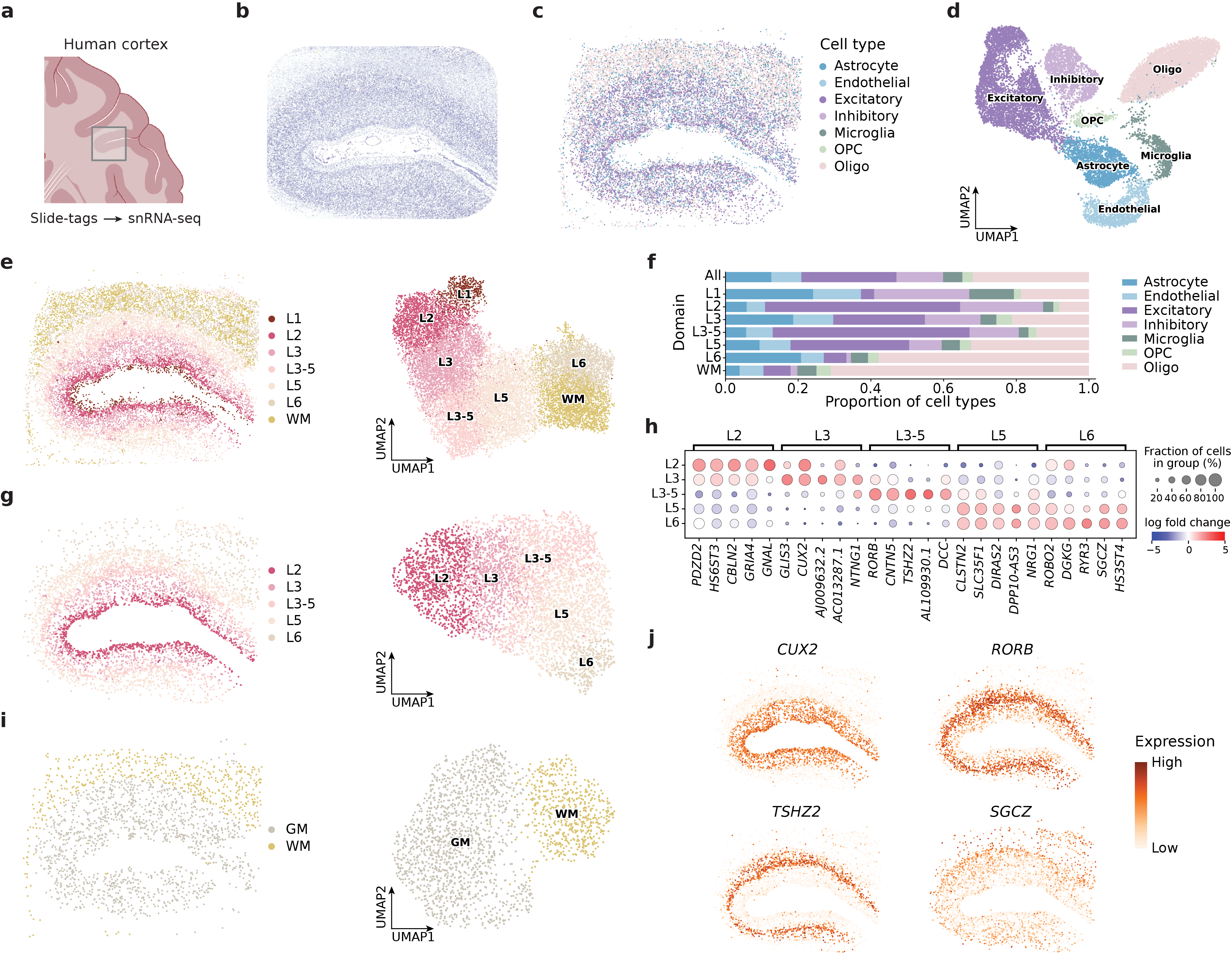}
\vspace{-.5em}
\caption{\textbf{CellScape maps spatial domains and cell types in the human cortex.} 
\textbf{a}, Spatial transcriptomics of human cortex profiled using Slide-tags, integrated with matched snRNA-seq.
\textbf{b}, Nissl-stained image of an adjacent tissue section.
\textbf{c}, Spatial distribution of cells colored by cell type annotations from snRNA-seq.
\textbf{d}, UMAP visualization of CellScape-learned intrinsic embeddings, colored by annotated cell types.
\textbf{e}, Spatial domains (left) and UMAP visualization of spatial embeddings (right) derived by CellScape; L1-6, cortical layers 1-6; WM, white matter.
\textbf{f}, Composition of major cell types across spatial domains identified by CellScape.
\textbf{g}, Spatial domains (left) and UMAP visualization of spatial embeddings (right) for excitatory neurons, both derived by CellScape.
\textbf{h}, Domain-specific marker gene expression across excitatory neuron subpopulations. Dot size indicates the fraction of expressing cells; color reflects log fold change.
\textbf{i}, Spatial domains (left) and UMAP visualization of spatial embeddings (right) for astrocytes, both derived by CellScape; GM, grey matter; WM, white matter.
\textbf{j}, Spatial distribution of representative domain-specific marker genes for excitatory neuron subpopulations identified in h.
}
\label{Fig:slidetags_human_cortex}
\vspace{-1em}
\end{figure}

To evaluate CellScape's efficacy in spatial tissue domain and cell type identification, we applied it to the Slide-tags human cortex dataset~\cite{russell2024slide}, which profiles 17,441 spatially resolved nuclei with transcriptome-wide expression at near-single-cell resolution across a prefrontal cortex section from a neurotypical donor~(Fig.~\ref{Fig:slidetags_human_cortex}a,b). CellScape-learned intrinsic embeddings \(( Z\_{\text{intrinsic}} \); see Methods) effectively recapitulated all major cell types (Fig.~\ref{Fig:slidetags_human_cortex}c,d), while spatial embeddings \( Z\_{\text{spatial}} \) sharply delineated cortical layers L1-6 and white matter (Fig.~\ref{Fig:slidetags_human_cortex}e). UMAP visualization of the spatial embeddings revealed an organization consistent with the known inside-out developmental layering of the cortex, progressing from white matter through deeper to superficial layers (Fig.~\ref{Fig:slidetags_human_cortex}e, right).
To further validate the biological coherence of the identified spatial domains, we analyzed the cell type composition with each region (Fig.~\ref{Fig:slidetags_human_cortex}f). Astrocytes were enriched in layer 1 and distributed across the grey matter; excitatory neurons peaked in layers 2–5, and inhibitory neurons were similarly confined to the grey matter. Oligodendrocytes (Oligos) dominated the white matter, consistent with their role in myelination~\cite{simons2016oligodendrocytes}, whereas microglia and oligodendrocyte precursor cells (OPCs) spanned both compartments. These distributions accord with established cortical architecture.

To investigate cell-type–specific spatial architecture, we applied CellScape separately to excitatory neurons and astrocytes. Within excitatory neurons (Fig.~\ref{Fig:slidetags_human_cortex}g), CellScape successfully identified canonical laminar markers such as \textit{CUX2} and \textit{RORB} (Fig.~\ref{Fig:slidetags_human_cortex}h) and demonstrated their expected layer-restricted expression patterns (Fig.~\ref{Fig:slidetags_human_cortex}j): \textit{CUX2} in L3, \textit{RORB} in L3–5, and \textit{SGCZ} in L6. In astrocytes, CellScape effectively distinguished protoplasmic astrocytes, enriched in grey matter, from fibrous astrocytes, predominantly located in white matter (Fig.~\ref{Fig:slidetags_human_cortex}i). Together, these results demonstrate CellScape’s capacity to detect fine-grained spatial heterogeneity both between and within cell types.

\subsection*{CellScape captures tissue and cellular-level variations within disease-associated domains}
\begin{figure}[!t]
\centering
    \includegraphics[width=0.95\linewidth]{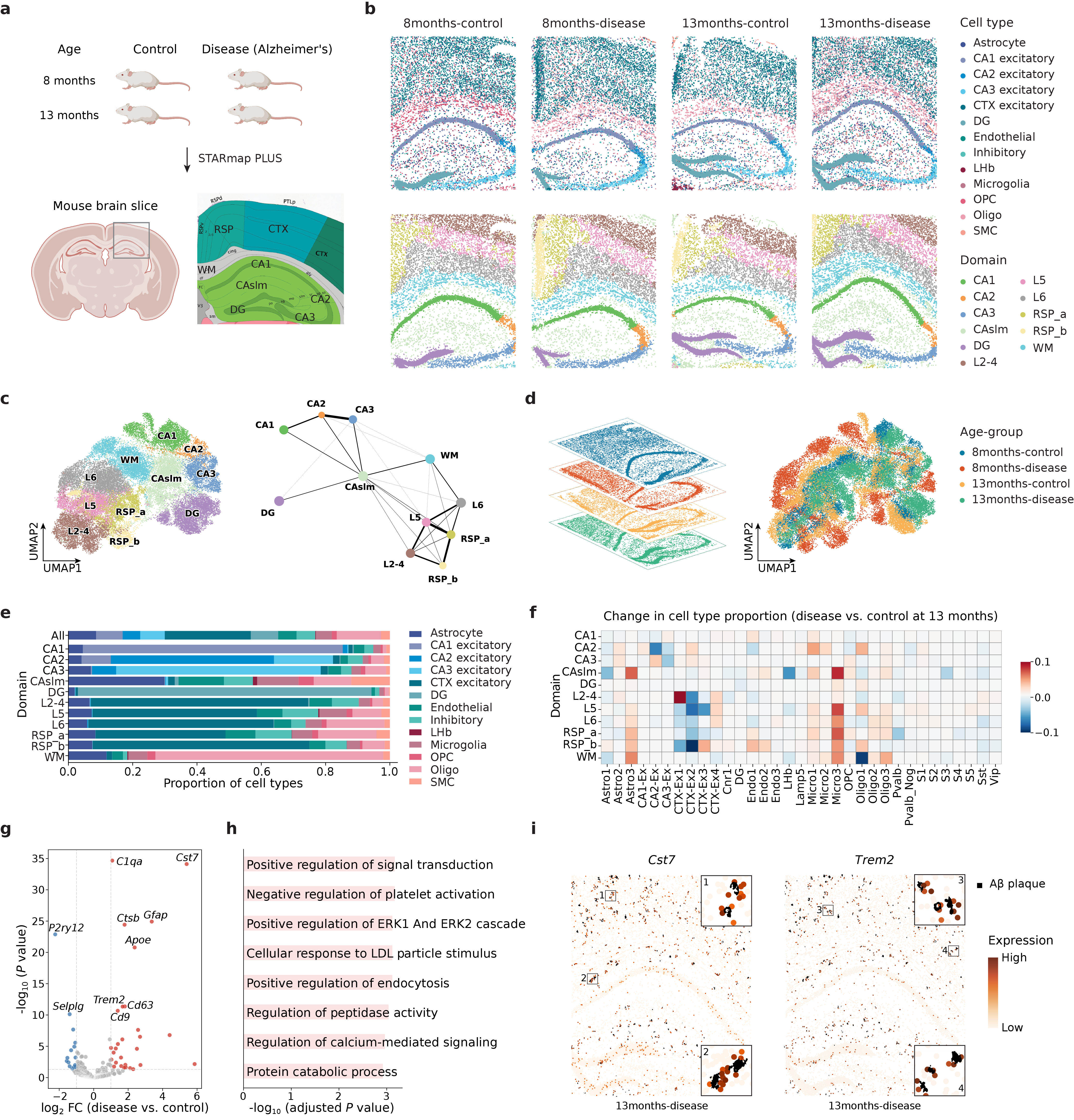}
\vspace{-.5em}
\caption{\textbf{CellScape reveals spatial and molecular alterations associated with Alzheimer’s disease in the mouse brain.} 
\textbf{a}, STARmap PLUS dataset contains coronal sections from Alzheimer’s disease (AD) mice and age-matched controls at 8 and 13 months. A reference image from the Allen Mouse Brain Atlas marks the anatomical region analyzed.
\textbf{b}, Spatial mapping of cells colored by cell types (top) and CellScape-defined spatial domains (bottom). CellScape consistently identifies anatomically coherent regions, including hippocampal subfields (CA1–CA3, CAslm), dentate gyrus (DG), cortical layers (L2–6), retrosplenial cortex regions (RSP-a, RSP-b), and white matter (WM).
\textbf{c}, UMAP visualization (left) and PAGA graph (right) of CellScape spatial embeddings.
\textbf{d}, UMAP of the four analyzed sections colored by sample.
\textbf{e}, Composition of major cell types across spatial domains.
\textbf{f}, Disease-associated shifts in cell-type composition at 13 months ($\Delta$= AD – control).
\textbf{g}, Volcano plot of genes differentially expressed in microglia (AD vs control, 13 months); red, up-regulated and blue, down-regulated in AD.
\textbf{h}, Gene Ontology enrichment for AD-up-regulated microglial genes.
\textbf{i}, Spatial maps of DAM markers \textit{Cst7} and \textit{Trem2} in a 13-month AD section; high expression colocalizes with $A\beta$ plaques (inset).
}
\label{Fig:starmap_mouse_brain}
\vspace{-1em}
\end{figure}
CellScape facilitates integrated spatial omics analysis across multiple samples. We applied CellScape to the STARmap PLUS mouse cortex dataset~\cite{zeng2023integrative}, encompassing subcellular in situ RNA imaging with multiplexed immunostaining for 2,766 genes across coronal sections from 8- and 13-month-old Alzheimer’s disease (AD) and control mice. Of the eight available sections, we analyzed four representative coronal slices (Fig.~\ref{Fig:starmap_mouse_brain}a,b). 
Joint embedding with CellScape resolved 11 transcriptionally distinct spatial domains (Fig.~\ref{Fig:starmap_mouse_brain}b), corresponding to hippocampal subfields (CA1–CA3), CA stratum lacunosum moleculare (CAslm), dentate gyrus (DG), cortical layers (L2–4, L5, L6), retrosplenial cortex regions (RSP-a, RSP-b), and white matter (WM) (Fig.~\ref{Fig:starmap_mouse_brain}c, left). The unified latent space preserved these anatomical distinctions (Fig.~\ref{Fig:starmap_mouse_brain}c, left) while separating subtly by genotype and age (Fig.~\ref{Fig:starmap_mouse_brain}d, right), indicating robust batch integration. Graph abstraction (PAGA)~\cite{wolf2019paga} on the CellScape clusters mirrored established circuit topology, with strong connections between contiguous hippocampal subfields and cortical layers, and CAslm positioned as a central node consistent with its role in cortico-hippocampal communication~\cite{maurin2014early} (Fig.~\ref{Fig:starmap_mouse_brain}c, right).
Domain-specific cell-type composition analysis reproduced canonical cytoarchitecture (Fig.~\ref{Fig:starmap_mouse_brain}e). Hippocampal fields were enriched for their cognate pyramidal neurons (CA1, CA2, CA3), dentate granule cells were confined to the dentate gyrus, and cortical/retrosplenial domains were dominated by CTX excitatory neurons. Oligos accumulated in white matter, whereas microglia were most abundant in the CAslm, consistent with single-cell brain atlases~\cite{zeisel2018molecular,zhang2023molecularly}.

To understand Alzheimer's-associated cellular remodeling, we compared 13-month AD sections with age-matched controls following multi-sample integration (Fig.~\ref{Fig:starmap_mouse_brain}f–h). CellScape revealed pronounced, domain-specific compositional shifts: a disease-enriched Micro3 microglial subset expanded in the stratum lacunosum-moleculare, upper cortical layers, and retrosplenial cortex (Fig.~\ref{Fig:starmap_mouse_brain}f). Differential expression between Micro3 and homeostatic microglia revealed the canonical disease-associated microglia (DAM) program (upregulation of \emph{Cst7}, \emph{Ctsb}, \emph{Apoe}, and \emph{Trem2}; downregulation of \emph{P2ry12} and \emph{Selplg}) (Fig.~\ref{Fig:starmap_mouse_brain}g). Gene-set enrichment highlighted pathways linked to microglial activation and lipid/immune signaling, including the ERK1/2 cascade, LDL-particle response, endocytosis, calcium signaling, and protein catabolism (Fig.~\ref{Fig:starmap_mouse_brain}h), indicating that microglia accumulating around $A\beta$ plaques are key drivers of the region-specific damage observed in Alzheimer’s disease~\cite{keren2017unique,choi2023autophagy,leng2021neuroinflammation}.
Lastly, we showed that regions with high expression of the DAM markers \emph{Cst7} and \emph{Trem2} colocalized with $A\beta$ plaques (Fig.~\ref{Fig:starmap_mouse_brain}i), underscoring the spatial coupling between plaque burden and microglial activation. 

\subsection*{CellScape reveals fine-grained tissue layers in the mouse olfactory bulb}
\begin{figure}[t!]
\centering
    \includegraphics[width=0.95\linewidth]{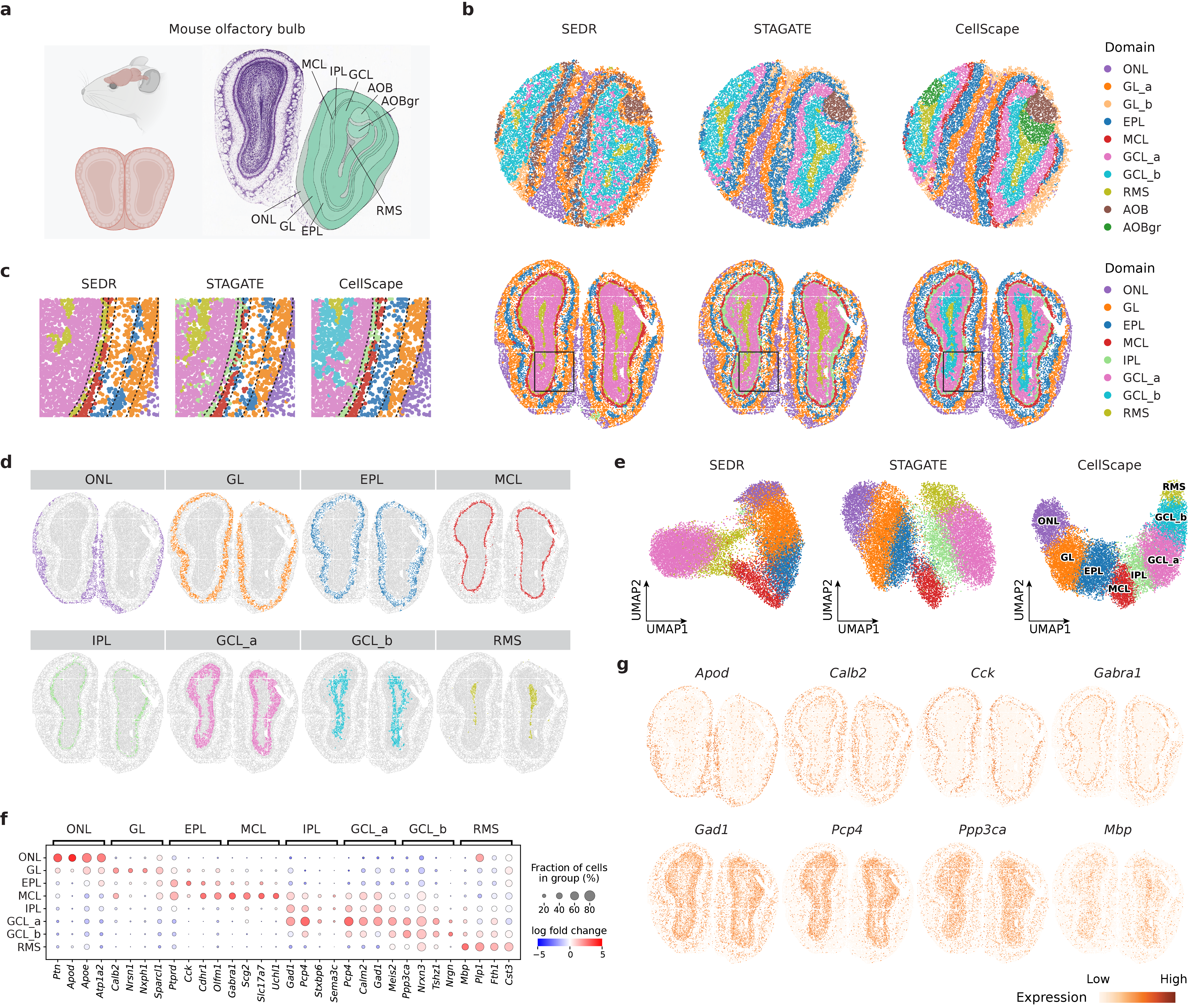}
\vspace{-.5em}
\caption{\textbf{CellScape identifies spatial domains and spatially variable genes for the mouse olfactory bulb.}
\textbf{a}, Mouse olfactory bulb profiled using Slide-seqV2 and Stereo-seq, with anatomical layers annotated based on the Allen Reference Atlas.
\textbf{b}, Spatial domain maps predicted by SEDR, STAGATE, and CellScape for Slide-seqV2 (top) and Stereo-seq (bottom). CellScape domains correspond to the olfactory nerve layer (ONL), glomerular layer (GL), external plexiform layer (EPL), mitral-cell layer (MCL), internal plexiform layer (IPL), granule-cell sublayers (GCL\_a, GCL\_b), rostral migratory stream (RMS), accessory olfactory bulb (AOB) and its granular layer (AOBgr).
\textbf{c}, Enlarged view of CellScape-predicted domains in Stereo-seq.
\textbf{d}, Individual display of each CellScape-identified domain in Stereo-seq.
\textbf{e}, UMAP visualization of embeddings from SEDR, STAGATE, and CellScape; only CellScape preserves the ordered layer continuum.
\textbf{f}, Domain-specific marker gene expression across CellScape domains (dot size, fraction of spots; color, log fold-change).
\textbf{g}, Spatial distribution of
representative marker genes associated with distinct olfactory bulb layers.
}
\label{Fig:mouse_olfactory_bulb}
\vspace{-1em}
\end{figure}

To assess spatial domain segmentation at near-single-cell resolution, we analyzed the mouse olfactory bulb data profiled with Slide-seqV2 (20,139 spots$\times$15,149 genes)~\cite{stickels2021highly} and Stereo-seq (19,109 spots$\times$14,376 genes)~\cite{chen2022spatiotemporal}. We benchmarked CellScape against graph-based methods SEDR~\cite{xu2024unsupervised} and STAGATE~\cite{dong2022deciphering}, using Allen Reference Atlas outlines to assign anatomical layer labels to each cluster (Fig.~\ref{Fig:mouse_olfactory_bulb}a). 
On Slide-seqV2, CellScape recovered all ten canonical layers, including the slender mitral-cell (MCL) and accessory-bulb granular (AOBgr) layers that the competing methods merged (Fig.~\ref{Fig:mouse_olfactory_bulb}b, top). In the Stereo-seq dataset, CellScape further separated the internal plexiform layer (IPL) and two granule-cell sublayers (GCL\_a, GCL\_b) that were less distinct in the segmentations generated by the competing methods (Fig.~\ref{Fig:mouse_olfactory_bulb}b-c). CellScape’s latent spatial embeddings preserved the expected trajectory from the rostral migratory stream (RMS) to the olfactory nerve layer (ONL), whereas those from SEDR and STAGATE appeared compressed (Fig.~\ref{Fig:mouse_olfactory_bulb}e). Finally, differential-expression analysis across CellScape domains recovered the canonical marker genes for each layer~\cite{xu2024unsupervised} (Fig.~\ref{Fig:mouse_olfactory_bulb}f,g), providing molecular validation of the segmentation.

\begin{figure}[!t]
\centering
    \includegraphics[width=0.9\linewidth]{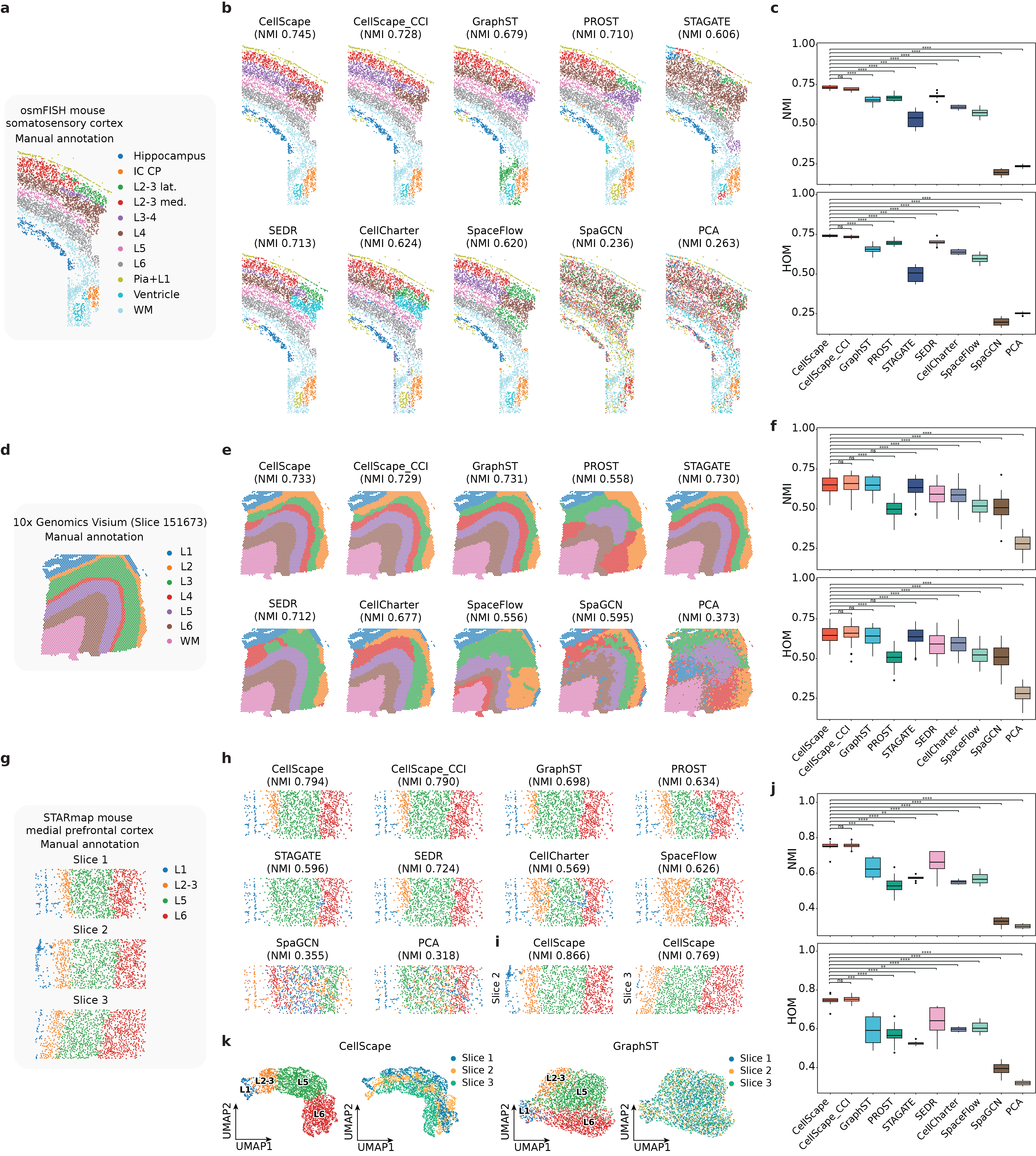}
\vspace{-.5em}
\caption{\textbf{CellScape accurately identifies spatial domains and enables multi-sample integration.}
\textbf{a}, Manual annotations for the osmFISH mouse somatosensory cortex dataset.
\textbf{b}, Predicted spatial domains for the osmFISH dataset by CellScape, CellScape\_CCI, and eight baseline methods. Cluster labels were matched to the closest reference domains.
\textbf{c}, Quantitative comparison of segmentation performance on the osmFISH dataset across ten runs with different random seeds. Box plots display the distribution of results (center line: mean; box limits: upper and lower quartiles; whiskers: 1.5$\times$ interquartile range).
\textbf{d}, Manual annotations for the 10x Genomics Visium human DLPFC dataset (Slice 151673).
\textbf{e}, Predicted spatial domains for Slice 151673 by each method.
\textbf{f}, Aggregated performance across all twelve DLPFC slices.
\textbf{g}, Manual annotations for the STARmap mouse medial prefrontal cortex dataset (three slices).
\textbf{h}, Predicted spatial domains for STARmap Slice 1 by each method.
\textbf{i}, Predicted spatial domains for STARmap Slices 2 and 3 by CellScape.
\textbf{j}, Quantitative evaluation of segmentation on STARmap Slice 1, summarized as in panel c. Statistical significance was assessed by two-sided t-test; **, ***, **** indicate $p < 0.01$, $< 0.001$, and $< 0.0001$, respectively.
\textbf{k}, UMAP visualization of cell embeddings learned by CellScape (left) and GraphST (right), colored by spatial domains and slice identity, respectively.
}
\label{Fig:benchmarking}
\vspace{-1em}
\end{figure}

\subsection*{CellScape improves spatial domain segmentation across diverse spatial omics datasets}
To systematically benchmark spatial domain segmentation, we evaluated CellScape across three spatial omics datasets with manual annotations: (1) the osmFISH mouse somatosensory cortex (SSC)  dataset~\cite{codeluppi2018spatial}, (2) the 10x Genomics Visium human dorsolateral prefrontal cortex (DLPFC) dataset~\cite{maynard2021transcriptome}, and (3) the STARmap mouse medial prefrontal cortex (mPFC) dataset~\cite{wang2018three}. Manual annotations served as ground truth, and segmentation accuracy was quantified using normalized mutual information (NMI) and homogeneity (HOM), two standard metrics for evaluating clustering consistency with reference labels~\cite{yuan2024benchmarking} (see Methods).

We compared CellScape against eight baseline methods: GraphST~\cite{long2023spatially}, PROST~\cite{liang2024prost}, STAGATE~\cite{dong2022deciphering}, SEDR~\cite{xu2024unsupervised}, CellCharter~\cite{varrone2024cellcharter}, SpaceFlow~\cite{ren2022identifying}, SpaGCN~\cite{hu2021spagcn}, and PCA followed by Louvain clustering, covering a range of spatial graph-based and non-spatial approaches.
For lower-resolution datasets such as Visium (55~$\mu$m spot diameter), where each spot aggregates transcripts from multiple cells, CellScape provides a spatial-only variant, CellScape\_CCI, which models spatial cellular interactions exclusively, as genomic relationships are less informative in mixed-cell profiles.

On the osmFISH SSC dataset, which profiles 33 genes at single-cell resolution across eleven annotated cortical regions (Fig.~\ref{Fig:benchmarking}a), CellScape achieved the highest segmentation accuracy, reaching a maximum NMI of 0.745 (Fig.~\ref{Fig:benchmarking}b). It accurately delineated most regions, with only minor mixing between L2–3 lat./med. and L3–4/L4. Across ten runs, CellScape consistently attained the highest mean NMI and HOM, with statistically significant improvements over all baselines (Fig.~\ref{Fig:benchmarking}c). Notably, CellScape outperformed its spatial-only variant (CellScape\_CCI) by 1\%, highlighting the advantage of incorporating genomic interactions in high-resolution datasets.

We next evaluated performance on the 10x Genomics Visium DLPFC dataset (55~$\mu$m resolution, twelve slices). On Slice 151673, a widely used ST benchmark (Fig.~\ref{Fig:benchmarking}d), CellScape achieved the highest NMI score of 0.733 (Fig.~\ref{Fig:benchmarking}e). Across all twelve slices (Fig.~\ref{Fig:benchmarking}f), CellScape (mean NMI 0.650, mean HOM 0.667) demonstrated performance statistically comparable to CellScape\_CCI (mean NMI 0.659, mean HOM 0.682), and significantly outperformed most other methods.
As expected, PCA with Louvain clustering, which does not incorporate spatial information, performed poorly (mean NMI 0.279, mean HOM 0.289), emphasizing the necessity of spatial modeling.

Finally, we applied CellScape to the STARmap mPFC dataset, which captures single-cell resolution profiles (Fig.~\ref{Fig:benchmarking}g). CellScape achieved the highest NMI (0.794) on Slice 1 (BZ9) (Fig.~\ref{Fig:benchmarking}h) and demonstrated significant improvements across multiple runs (Fig.~\ref{Fig:benchmarking}j). CellScape further enabled integrated segmentation across slices, maintaining high accuracy on Slices 2 and 3 (Fig.~\ref{Fig:benchmarking}i). UMAP visualization of CellScape-integrated embeddings preserved the laminar organization from L1 to L6 while maintaining slice-specific distinctions, outperforming GraphST, which exhibited less organized spatial separation and blurred layer transitions (Fig.~\ref{Fig:benchmarking}k).


\section*{Discussion}
Tissue organization is fundamentally governed by two distinct yet intertwined forms of interaction. These include extrinsic (intercellular) interactions, such as spatial proximity and cell--cell communication, and intrinsic (intracellular) interactions, such as regulatory gene--gene relationships or gene co-expression networks within individual cells. While these two types of interactions reflect different aspects of biological regulation, their influences on tissue organization are deeply interwoven and jointly determine cellular identity and function. Consequently, the effective modeling of tissue architecture
and cellular phenotypes requires the explicit integration of both the spatial context and the intracellular regulatory landscape.

A range of computational methods have been developed for ST analysis, aiming to model tissue organization and identify spatially informed cellular phenotypes. Among these, graph learning has been a predominant paradigm, where cells represented as nodes and spatial proximities as edges within the modeling framework~\cite{tie2022contextual,li2022graph}. Most of the approaches~\cite{long2023spatially,liang2024prost,dong2022deciphering,xu2024unsupervised,ren2022identifying,hu2021spagcn} rely on GNNs to propagate information across neighborhoods. However, message passing alone leaves the intrinsic gene–gene regulatory structure implicit. Furthermore, GNNs are prone to oversmoothing in deeper networks~\cite{li2018deeper}, limiting their ability to fully recovercoupled spatial and transcriptional signals that are crucial for comprehensive ST analysis.

CellScape provides an effective tool for the autodelineation of the spatial and functional landscape of complex tissues. The spatial branch models intercellular interactions by applying contrastive learning to capture neighborhood relationships. Meanwhile, gene-gene relationships and regulatory structure are extracted by CNN analysis of 2D genomic maps~\cite{yan2024interpretable} of the gene expression data. The outputs of the two branches are integrated by a shared embedding layer, producing unified cell representations that simultaneously capture transcriptional state, spatial neighborhood, and regulatory structure.
The proposed CellScape provides a richer and more disentangled characterization of cellular identity and tissue organization. Additionally, CellScape incorporates integrated batch correction for multi-sample consistency and offers a spatial-only variant (CellScape\_CCI) for spot-based ST platforms where genomic interaction modeling is less informative.


We benchmarked CellScape across diverse ST datasets, demonstrating its strong performance across multiple downstream applications. It consistently outperformed state-of-the-art methods in spatial domain segmentation (Figs.~\ref{Fig:mouse_olfactory_bulb} and \ref{Fig:benchmarking}) and cell type identification (Figs.~\ref{Fig:slidetags_human_cortex}c). Importantly, in multi-sample analyses, CellScape effectively corrected for batch effects while preserving biologically relevant variations, thereby enabling integrated analyses of tissue architecture and cellular composition across different conditions (Fig.~\ref{Fig:starmap_mouse_brain}). Beyond spatial domain segmentation, CellScape supports a range of downstream tasks, including spatial domain transition analysis, domain-specific marker gene discovery, functional enrichment analysis, and tissue-scale cell type composition analysis. This broad utility underscores its versatility in resolving complex spatial and molecular patterns.
The dual-interaction learning strategy underlying CellScape is broadly applicable to other domains where both entity-level and feature-level interactions are critical for capturing system structure.
For instance, in scRNA-seq, cell–cell graphs can be constructed from transcriptomic similarity or inferred ligand–receptor interactions, enabling the integration of intercellular communication with gene regulatory programs. 
The framework naturally extends to spatial epigenomics~\cite{deng2022spatial}, cancer genomics~\cite{knoche2021role}, and neuroscience~\cite{lynn2019physics}, where tissue structure, tumor–microenvironment interactions, and regional connectivity, respectively, complement molecular signals. 
Beyond biomedicine, dual-interaction modeling is also applicable to recommender systems and tabular data, where capturing inter-entity and inter-feature dependencies improves representation learning. 
These diverse applications underscore the versatility of CellScape’s modular design across biological and computational domains.

In summary, CellScape provides a flexible framework for ST analysis to simultaneously consider spatial organization and gene regulatory interactions. As spatial technologies continue to evolve, such integrative analysis is becoming essential for advancing the field, offering critical insights into tissue architecture, development, and disease. Beyond spatial omics, CellScape's dual-interaction design provides a general strategy for multi-level representation learning, boasting broad applicability in diverse domains where structured relationships among entities and features are fundamental.

\section*{Methods}
Fig.~\ref{Fig:framework}a outlines the CellScape workflow, a unified framework that learns cell representations by integrating spatial organization and gene co-expression patterns in ST data. Starting from spatial coordinates of individual cells and a gene expression matrix, CellScape derives a cell–cell spatial proximity matrix and a gene–gene co-expression matrix. The spatial matrix is used to build a cell graph, while the gene matrix is embedded into a 2D expression map per cell, where proximally co-expressed genes are placed closer together and pixel intensities encode expression levels. Through self-supervised learning with contrastive and reconstruction objectives, CellScape jointly captures intercellular spatial context and intracellular gene interactions to derive informative cell embeddings. These representations enable downstream analyses such as spatial domain segmentation (Fig.~\ref{Fig:framework}b).

\subsection*{CellScape}
\subsubsection*{Preliminaries}

Consider a ST dataset consisting of a gene expression matrix
$
X = [\mathbf{x}^{(1)}, \mathbf{x}^{(2)}, \dots, \mathbf{x}^{(n)}] \in \mathbb{R}^{p \times n},
$
where each column \( \mathbf{x}^{(i)} \in \mathbb{R}^p \) represents the expression profile of \( p \) genes for cell \( i \), and \( n \) is the total number of cells.
The corresponding spatial coordinates are given by
$
\mathcal{C} = [\mathbf{c}^{(1)}, \mathbf{c}^{(2)}, \dots, \mathbf{c}^{(n)}] \in \mathbb{R}^{2 \times n},
$
where each column \( \mathbf{c}^{(i)} \in \mathbb{R}^2 \) denotes the spatial location (e.g., \( x \)- and \( y \)-coordinates) of cell \( i \) on the tissue.
CellScape learns two complementary low-dimensional embeddings for each cell: a spatial embedding $Z_{\text{spatial}}$ that captures spatial organization and an intrinsic embedding $Z_{\text{intrinsic}}$ that captures cell-intrinsic transcriptional structure. These embeddings are utilized for downstream analyses.


\subsubsection*{Spatial graph construction for cellular interaction modeling}

To model spatial interactions among cells, we construct a cell--cell interaction matrix \( A \in \mathbb{R}^{n \times n} \), where edge weights reflect spatial proximity based on the coordinates \( \mathcal{C} \). For Visium and Stereo-seq datasets, which exhibit relatively uniform and grid-like structures, we define edges using a \textit{k}-nearest neighbors (kNN) approach, connecting each cell to its \( k \) closest neighbors based on Euclidean distance. 
For datasets with irregular spatial layouts, such as Slide-seqV2, seqFISH, and STARmap, we define edges using Delaunay triangulation~\cite{hagberg2008exploring}, which adapts to local spatial density and better preserves topological continuity in heterogeneous tissues.
In both cases, the adjacency matrix is defined as:
$$
A_{ij} =
\begin{cases}
\left\| \mathbf{c}^{(i)} - \mathbf{c}^{(j)} \right\|_2, & \text{if } (i, j) \in \mathcal{E}, \\
0, & \text{otherwise},
\end{cases}
$$
where \( \mathcal{E} \subseteq \{(i, j) \mid i \neq j\} \) denotes the set of connected cell pairs and \( \| \cdot \|_2 \) denotes the Euclidean norm.\\

\noindent Based on the cell--cell interaction matrix \( A \), we construct an undirected graph \( G = (V, E) \), where each node \( v_i \in V \) represents cell \( i \), and an edge \( (v_i, v_j) \in E \) exists if and only if \( A_{ij} > 0 \). The corresponding edge weight is given by the spatial proximity value \( A_{ij} \). This graph captures the tissue’s spatial organization and serves as the structural backbone for spatial modeling.

\subsubsection*{Gene expression map construction for genomic interaction modeling}
To model gene--gene interactions within each cell, we compute a gene co-expression matrix \( C \in \mathbb{R}^{p \times p} \), where each entry is defined as
$
C_{uv} = \mathrm{corr}(X_{u \cdot}, X_{v \cdot}),
$
with \( \mathrm{corr}(\cdot, \cdot) \) denoting the Pearson correlation between the expression profiles of genes \( u \) and \( v \) across all cells.
To encode this information, we transform each 1D expression vector \( \mathbf{x}^{(i)} \in \mathbb{R}^p \) into a 2D matrix \( M^{(i)} \in \mathbb{R}^{q \times q} \), where \( q = \lceil \sqrt{p} \rceil \). If \( p < q^2 \), the remaining \( q^2 - p \) entries are padded with zeros.
This transformation~\cite{yan2024interpretable} is guided by an optimal transport-based algorithm applied to the gene co-expression matrix \( C \), which assigns genes to grid positions such that highly co-expressed genes are placed close together. Let \( \pi: \{1, \dots, p\} \rightarrow \{1, \dots, q\} \times \{1, \dots, q\} \) denote the learned spatial assignment of genes. The resulting 2D map is defined as:
$$
M^{(i)}_{r,s} =
\begin{cases}
x^{(i)}_{\pi^{-1}(r, s)}, & \text{if } (r, s) \in \text{grid}(\pi), \\
0, & \text{otherwise},
\end{cases}
$$
where \( \pi^{-1}(r, s) \) returns the gene index assigned to grid position \( (r, s) \). This 2D layout preserves the gene co-expression pattern and enables convolutional neural networks to effectively capture genomic interactions within each cell.

\subsubsection*{Dual-interaction representation learning}
We introduce a dual-interaction representation learning model to capture both the spatial context and intrinsic transcriptional patterns of individual cells. CellScape learns two complementary embeddings: a spatial embedding reflecting spatial proximity among cells within the tissue and an intrinsic embedding capturing gene co-expression patterns within each cell. The spatial embedding is learned through contrastive learning applied to neighborhoods sampled from a spatial cell--cell graph, while the intrinsic embedding is derived from the masked reconstruction of 2D gene expression maps.

\paragraph{Feature masking.}
To enhance the robustness and context-awareness of learned representations, we adopt a masking strategy inspired by masked autoencoding
~\cite{he2022masked, hou2022graphmae}. 
For a randomly selected subset of cells \( \mathcal{M} \subset V \), we simultaneously mask both the feature vector \( \mathbf{x}^{(i)} \in \mathbb{R}^p \) and its corresponding gene expression map \( M^{(i)} \in \mathbb{R}^{q \times q} \). The masked inputs are defined as:
$$
\tilde{\mathbf{x}}^{(i)} =
\begin{cases}
\mathbf{0}, & \text{if } i \in \mathcal{M}, \\
\mathbf{x}^{(i)}, & \text{otherwise},
\end{cases}
\quad
\tilde{M}^{(i)} =
\begin{cases}
\mathbf{0}, & \text{if } i \in \mathcal{M}, \\
M^{(i)}, & \text{otherwise}.
\end{cases}
$$
The masked features \( \tilde{\mathbf{x}}^{(i)} \) are passed into a Graph Attention Network (GAT) encoder over the spatial graph, and the masked image \( \tilde{M}^{(i)} \) is processed by a Convolutional Neural Network (CNN) encoder. The outputs of these encoders are then concatenated and projected into a shared latent representation \( \mathbf{z}^{(i)} \). To reconstruct the original features, we apply a GAT decoder:
$$
\hat{\mathbf{x}}^{(i)} = f_{\text{DEC}}(G, \mathbf{z}^{(i)}).
$$
The reconstruction is supervised by a scaled cosine error (SCE) loss:
$$
\mathcal{L}_{\text{recon}} = \frac{1}{|\mathcal{M}|} \sum_{i \in \mathcal{M}} \left(1 - \cos(\mathbf{x}^{(i)}, \hat{\mathbf{x}}^{(i)}) \right)^\gamma,
$$
where \( \cos(\cdot, \cdot) \) denotes cosine similarity, and \( \gamma \) is a scaling exponent used to amplify larger angular errors. We set \( \gamma = 3 \) to penalize inaccurate reconstructions more strongly and encourage the model to learn precise and discriminative representations. This masking strategy encourages the model to reconstruct missing information by leveraging both spatial and transcriptional context, thus deriving generalizable and biologically meaningful representations.

\paragraph{Neighborhood sampling.}
While feature masking and reconstruction guide the model to recover missing information, it does not explicitly capture the tissue spatial structure. 
To model local cellular environments, we introduce a contrastive learning objective that encourages cells in spatial proximity to have similar embeddings. This is motivated by the biological observation that neighboring cells within the same microenvironment often exhibit shared functional behaviors or engage in coordinated biological processes. By learning embeddings that reflect this spatial organization, CellScape aims to capture crucial aspects of cellular spatial context and interaction.

For each cell \(i\), we define a set of spatial neighbors \(\mathcal{N}_i\) derived from a spatial graph, and treat their embeddings as positive samples relative to the anchor cell \(i\). To learn spatially coherent representations, we adopt a batch-wise multi-positive contrastive loss inspired by MIL-NCE~\cite{miech2020end}, which simultaneously attracts embeddings of neighboring cells while repelling those of unrelated cells sampled from other parts of the tissue. The loss is formulated as:
$$
\mathcal{L}_{\text{contrastive}}
=
\frac{1}{n}
\sum_{i=1}^n
- \log
\frac{\sum_{j \in \mathcal{N}_i}\exp\left(\mathbf{z}^{(i)} \cdot \mathbf{z}^{(j)} / \tau\right)}
{\sum_{k=1}^n \sum_{l \in \mathcal{N}_k}\exp\left(\mathbf{z}^{(i)} \cdot \mathbf{z}^{(l)} / \tau\right)},
$$
where \( \mathbf{z}^{(i)} \) is the \(\ell_2\)-normalized embedding of cell \(i\), and \( \tau \) is a temperature parameter that modulates the separation between positive and negative samples. The numerator aggregates similarities between the anchor and its immediate neighbors, while the denominator pools similarities across all neighbor sets in the batch, implicitly treating non-neighboring cells as negatives. This loss promotes the emergence of a latent space in which spatially adjacent cells cluster together, enhancing the model’s ability to reflect biologically meaningful spatial organization. When combined with feature reconstruction, this spatial constraint enables CellScape to learn representations that integrate both local cellular context and transcriptomic identity.

\paragraph{Model architecture.}
CellScape employs a dual-encoder architecture to learn complementary representations that integrate spatial and transcriptional information for each cell.

To capture spatial context, a multi-layer GAT encoder processes the masked node features \( \tilde{\mathbf{x}}^{(i)} \) defined on the spatial graph \( G \), yielding spatial embeddings \( \mathbf{z}_{\text{spatial}}^{(i)} \). The GAT encoder consists of multiple stacked graph attention layers. Each layer applies a multi-head attention mechanism to aggregate and transform node features based on their neighbors. Specifically, at each layer, the embedding of a node \( i \) (\( \mathbf{h}_i^{(l)} \) at layer \( l \)) is updated by attending to the features of its neighbors \( \mathcal{N}_i \), where the attention weights \( \alpha_{ij} \) are learned adaptively based on an attention mechanism \( \alpha_{\text{attn}} \):
$$
e_{ij} = \alpha_{\text{attn}}\left(W^{(l)}\mathbf{h}_i^{(l)},\, W^{(l)}\mathbf{h}_j^{(l)}\right), \quad
\alpha_{ij} = \frac{\exp(e_{ij})}{\sum_{k\in \mathcal{N}_i}\exp(e_{ik})},
$$
where \( W^{(l)} \) is a learnable weight matrix. This process allows the network to weigh the importance of different neighbors based on their relevance to the central node. The output of the final GAT layer provides the spatial embedding \( \mathbf{z}_{\text{spatial}}^{(i)} \).

In parallel, to extract intrinsic transcriptional patterns, a CNN encoder processes the masked 2D gene expression map \( \tilde{M}^{(i)} \). The CNN encoder comprises a sequence of convolutional layers with non-linear activations and batch normalization to capture local co-expression patterns within the gene expression map. The final feature maps are then flattened and projected through a fully connected layer to produce the intrinsic embedding \( \mathbf{z}_{\text{intrinsic}}^{(i)} \), which encapsulates the key transcriptional relationships encoded in the spatial arrangement of the gene expression map:
$$
\mathbf{z}_{\text{intrinsic}}^{(i)} = \mathrm{FC}(\mathrm{Flatten}(\mathrm{Conv2D}(\tilde{M}^{(i)}))).
$$

Finally, the learned spatial embedding \( \mathbf{z}_{\text{spatial}}^{(i)} \) and the intrinsic embedding \( \mathbf{z}_{\text{intrinsic}}^{(i)} \) are concatenated to form a joint embedding that integrates both views of cellular information. This combined representation is then linearly projected through a learnable weight matrix \( W \) to yield the unified latent embedding \( \mathbf{z}^{(i)} \) for each cell:
$$
\mathbf{z}^{(i)} = W \left[\mathbf{z}_{\text{spatial}}^{(i)} \| \mathbf{z}_{\text{intrinsic}}^{(i)}\right].
$$
This fused embedding is used to jointly optimize the parameters of both the GAT and CNN encoders through the feature reconstruction loss and the contrastive structural loss. Following training, the spatial embeddings \( \mathbf{z}_{\text{spatial}}^{(i)} \) and the intrinsic embeddings \( \mathbf{z}_{\text{intrinsic}}^{(i)} \) can be independently extracted from their respective encoders, capturing distinct perspectives of cellular information. The spatial embeddings are well-suited for tasks such as spatial domain segmentation, and the intrinsic embeddings can be effectively used for cell type clustering. Furthermore, these two complementary embeddings can also be combined or used in a multi-view learning framework for more comprehensive downstream analyses.



\paragraph*{Model training.}
The model is trained by jointly considering a feature reconstruction loss (\( \mathcal{L}_{\text{intrinsic}} \)) for learning intrinsic transcriptional patterns and a contrastive loss (\( \mathcal{L}_{\text{spatial}} \)) for learning spatial context. To manage potential conflicts between these objectives, we employ projected conflict gradient (PCGrad)~\cite{yu2020gradient}, which adaptively adjusts the gradients of each loss to mitigate negative interference during backpropagation. This promotes stable joint learning of both intrinsic and spatial representations.
Model parameters are optimized using the Adam optimizer with an initial learning rate of \( 1 \times 10^{-3} \) and a weight decay of \( 1 \times 10^{-4} \). A learning rate scheduler reduces the learning rate by a factor of 0.5 every 50 epochs to facilitate convergence. 

\subsubsection*{Batch effect correction for multi-sample analysis}
To enable integrated analysis across multiple samples, we corrected batch effects at both the molecular and spatial levels. Gene expression was first harmonized using ComBat~\cite{johnson2007adjusting}, implemented via the \texttt{sc.pp.combat} function in Scanpy with the appropriate batch key, to remove batch-specific biases while preserving biological variability.
To address spatial-level batch effects, we constructed spatial graphs independently for each sample based on the spatial coordinates of cells. The resulting sample-specific adjacency matrices were then combined into a global block-diagonal graph, preserving intra-sample spatial relationships while avoiding artificial inter-sample connections.





\subsection*{Downstream analyses}
\subsubsection*{Spatial domain segmentation}
Spatial domains are tissue regions composed of transcriptionally and spatially coherent cell populations, often corresponding to distinct anatomical or functional structures~\cite{staahl2016visualization,maynard2021transcriptome}. To identify these domains, we perform unsupervised clustering on low-dimensional spatial embeddings of cells learned by CellScape.
We first apply principal component analysis (PCA) to the learned spatial embeddings \( Z_{\text{spatial}} \in \mathbb{R}^{n \times d} \), projecting them into \( Z_{\text{spatial\_PCA}} \in \mathbb{R}^{n \times k} \) (typically \( k = 30 \)) while retaining the most informative variance.
The reduced embeddings are then clustered using a Gaussian mixture model implemented via the \texttt{mclust} algorithm~\cite{scrucca2023model}. Each cell is assigned to a cluster based on the maximum posterior probability.
To enhance spatial continuity, we optionally refine cluster assignments via majority voting among spatial neighbors. Specifically, each cell's label is updated to match the most frequent cluster label among its \( r \) nearest neighbors in Euclidean space.

The performance of our spatial domain segmentation was quantitatively evaluated using normalized mutual information (NMI) and homogeneity (HOM)~\cite{rosenberg2007v}, established metrics for evaluating the consistency between unsupervised clustering results and reference annotations. 

NMI has been widely adopted for benchmarking spatial domain segmentation methods~\cite{yuan2024benchmarking}, as it measures the agreement between predicted clusters \( \hat{Y} \) and reference annotations \( Y \). NMI is defined as:
$$
\mathrm{NMI}(Y, \hat{Y}) = \frac{2 \cdot I(Y; \hat{Y})}{H(Y) + H(\hat{Y})},
$$
where \( I(Y; \hat{Y}) \) is the mutual information between the predicted and true labels, and \( H(\cdot) \) denotes entropy.

HOM score quantifies the purity of the identified spatial domains by measuring the degree to which each predicted domain contains cells exclusively from a single ground truth tissue region:
$$
\mathrm{HOM}(Y, \hat{Y}) = 1 - \frac{H(Y \mid \hat{Y})}{H(Y)},
$$
where $( H(Y \mid \hat{Y}) )$ is the conditional entropy of the ground truth labels given the predicted domains. Both metrics range from 0 to 1, with higher values indicating stronger concordance between the identified spatial domains and the underlying tissue structure.

\subsubsection*{Spatial domain transition}
To infer potential developmental trajectories or transitions between spatial domains, we applied partition-based graph abstraction (PAGA)~\cite{wolf2019paga} to the spatial embeddings \( Z_{\text{spatial}} \) learned by CellScape. PAGA constructs a coarse-grained graph that preserves the topological structure of the data by connecting transcriptionally similar domains based on shared nearest neighbors. In this graph, nodes represent spatial domains and edge weights reflect the degree of connectivity, providing a high-level view of potential lineage relationships and gradual transitions in cellular states across the tissue.


\subsubsection*{Spatial domain-specific marker gene identification}
To characterize the transcriptional programs of each spatial domain identified by CellScape, we performed differential gene expression (DGE) analysis using the Wilcoxon rank-sum test, implemented via the \texttt{sc.tl.rank\_genes\_groups} function in Scanpy. Differentially expressed genes (DEGs) were identified for each domain by comparing expression levels against all other domains, based on adjusted P-values and fold changes.
DEGs were further filtered using a minimum fold-change threshold, and top-ranked, domain-specific genes were selected as candidate marker genes. These markers defined the transcriptional signatures of each domain and were visualized using dot plots summarizing expression levels and the fraction of expressing cells across domains.

\subsubsection*{Spatial domain functional enrichment analysis}
To elucidate the functional roles of each spatial domain, we performed gene set enrichment analysis on domain-specific marker genes using the Enrichr module in the GSEApy package\cite{fang2023gseapy,kuleshov2016enrichr}. Genes were tested for enrichment against curated gene sets, including GO Biological Process (GOBP)~\cite{gene2023gene} and KEGG pathways, with significance defined as adjusted $P<0.05$. The analysis presented in Fig.~\ref{Fig:starmap_mouse_brain}h focused on the GOBP gene set. Top enriched terms were ranked by significance and visualized as bar plots, revealing distinct biological processes associated with each spatial domain.

\subsubsection*{Tissue-scale compositional analysis}
To investigate the spatial organization of cellular populations, we performed a tissue-scale compositional analysis by quantifying the distribution of annotated cell types across the identified spatial domains. For each domain \( i \), we computed the proportion \( P_{ij} \) of each cell type \( j \) as the number of cells of type \( j \) within domain \( i \) (\( N_{ij} \)) normalized by the total number of cells in that domain:
$$
P_{ij} = \frac{N_{ij}}{\sum_{j} N_{ij}}.
$$
This yielded a domain-specific cell type composition matrix \( P \), with each row representing the normalized cell type composition of a given domain. As a reference, we also computed the overall proportion of each cell type across the entire tissue:
$$
P_{\text{all}, j} = \frac{\sum_i N_{ij}}{\sum_{i,j} N_{ij}}.
$$
The matrix \( P \) and the global proportions \( P_{\text{all}} \) revealed spatial heterogeneity in cellular composition, exemplified by stacked bar plots in Fig.~\ref{Fig:slidetags_human_cortex}h and Fig.~\ref{Fig:starmap_mouse_brain}e, which highlight domain-enriched cell types and compositional shifts across the tissue. This approach links cell type organization to spatial context and enables cross-sample comparisons to identify domain- and cell type-specific changes under different biological conditions, providing a framework for studying spatial tissue remodeling and underlying cellular dynamics, as demonstrated in Fig.~\ref{Fig:starmap_mouse_brain}f.



\subsection*{Method benchmarking for spatial domain segmentation}

We systematically benchmarked CellScape against eight representative ST representation learning methods: GraphST~\cite{long2023spatially}, PROST~\cite{liang2024prost}, STAGATE~\cite{dong2022deciphering}, SEDR~\cite{xu2024unsupervised}, CellCharter~\cite{varrone2024cellcharter}, SpaceFlow~\cite{ren2022identifying}, SpaGCN~\cite{hu2021spagcn}, and PCA. All methods were evaluated using a standardized three-stage pipeline comprising data preprocessing, representation learning, and spatial clustering.

\subsubsection*{Data preprocessing}
Raw gene expression counts were normalized to 10,000 transcripts per cell using \texttt{sc.pp.normalize\_total} and log-transformed with \texttt{sc.pp.log1p} in Scanpy. The top 3,000 highly variable genes (HVGs) were identified from the raw counts using Seurat\_v3~\cite{stuart2019comprehensive} via \texttt{sc.pp.highly\_variable\_genes}, selecting genes with the greatest variability across cells. The resulting HVG expression profiles were used as input for model training across all methods for benchmarking.

\subsubsection*{Representation learning across methods}

The representation learning strategy varied across methods. GraphST\cite{long2023spatially} uses graph attention networks to model local spatial dependencies and gene expression patterns. PROST\cite{liang2024prost} applies spatial graph contrastive learning to generate context-aware embeddings. STAGATE\cite{dong2022deciphering} leverages graph attention to integrate spatial structure with transcriptomic profiles. SEDR\cite{xu2024unsupervised} encodes cells using a deep autoencoder combined with a spatial constraint on a cell graph. CellCharter\cite{varrone2024cellcharter} adopts a variational autoencoder framework with spatial neighbor aggregation for representation learning. SpaceFlow\cite{ren2022identifying} combines graph contrastive learning with spatial regularization to derive spatially coherent embeddings. SpaGCN\cite{hu2021spagcn} integrates spatial coordinates and gene expression via graph convolutional networks. PCA serves as a baseline, performing linear dimensionality reduction using only gene expression data.

\subsubsection*{Spatial clustering}
Spatial clustering was performed on the learned representations from each method, utilizing the clustering algorithms originally proposed for each: Gaussian mixture modeling for CellScape, CellCharter, STAGATE, SEDR, GraphST, and PROST; Louvain clustering for SpaGCN and PCA; and Leiden algorithm for SpaceFlow. Method-specific refinement was applied when available; otherwise, spatial domain labels were refined using spatial smoothing based on local majority voting. Clustering performance was assessed using NMI and HOM against expert-annotated labels. 


\subsection*{Datasets}

\subsubsection*{Slide-tags human cortex data} The human prefrontal cortex dataset generated using the Slide-tags technology~\cite{russell2024slide} was obtained from \url{https://singlecell.broadinstitute.org/single_cell/study/SCP2167}. Slide-tags integrates spatial indexing with snRNA-seq by capturing nuclei from a 20 $\mu$m thick fresh frozen tissue section onto a monolayer of spatially barcoded beads. 
This sequencing-based method achieves single-cell resolution across a 100 mm$^2$ region of the prefrontal cortex from a neurotypical 78-year-old donor. The dataset contains 17,441 spatially mapped nuclei with a median of 3,196 unique molecular identifiers (UMIs) per nucleus, with gene expression measured transcriptome wide.

\subsubsection*{STARmap PLUS mouse brain data} The mouse cortex dataset generated using the STARmap PLUS technology~\cite{zeng2023integrative} was obtained from \url{https://singlecell.broadinstitute.org/single_cell/study/SCP1375}. STARmap PLUS is an imaging-based in situ ST method that simultaneously profiles targeted mRNA expression and protein localization at subcellular resolution (95×95×350 nm voxels). In this study, coronal brain sections were collected from 8- and 13-month-old TauPS2APP transgenic mice, an established model of Alzheimer's disease characterized by amyloid plaque and tau pathology, and from age-matched control mice. Gene expression was measured for a targeted panel of 2,766 genes across eight slices, with 35,094 cells retained from four selected samples for downstream analysis.

\subsubsection*{Slide-seqV2 mouse olfactory bulb data} The Slide-seqV2 mouse olfactory bulb dataset~\cite{stickels2021highly} was obtained from \url{https://singlecell.broadinstitute.org/single_cell/study/SCP815} (Puck\_200127\_15). Slide-seqV2 is a bead-based ST technology that captures mRNA expression at near-single-cell resolution using 10 $\mu$m spatially barcoded beads. This dataset profiles 20,139 spatial spots and 21,220 genes across a sagittal section of the olfactory bulb.

\subsubsection*{Stereo-seq mouse olfactory bulb data} The Stereo-seq mouse olfactory bulb dataset~\cite{chen2022spatiotemporal} was obtained from \url{https://github.com/JinmiaoChenLab/SEDR_analyses/tree/master/data}.  
Stereo-seq is a DNA nanoball-based ST platform that achieves subcellular resolution, with individual DNA nanoballs spaced $\sim$500 nm apart. For analysis, the original data were aggregated into non-overlapping bins of 14$\times$14 nanoballs, yielding an effective resolution of approximately 10 $\mu$m per spot. 
This dataset profiles 19,109 spatial spots and 27,106 genes across an olfactory bulb section.

\subsubsection*{osmFISH mouse somatosensory cortex data} The osmFISH mouse somatosensory (SS) cortex dataset~\cite{codeluppi2018spatial} was obtained from \url{https://linnarssonlab.org/osmFISH/}. osmFISH is an imaging-based method that detects RNA molecules at single-molecule resolution. This dataset profiles 1,976,659 RNA molecules across 4,839 segmented cells, measuring expression for a targeted panel of 33 genes.

\subsubsection*{10x Genomics Visium human DLPFC data} The human dorsolateral prefrontal cortex (DLPFC) dataset generated using the 10x Genomics Visium platform~\cite{maynard2021transcriptome} was obtained from \url{https://github.com/LieberInstitute/spatialLIBD}. This ST dataset profiles gene expression across twelve tissue sections collected from three neurotypical adult donors. Gene expression measurements cover 33,538 genes across spatial spots at 55 $\mu$m resolution, where each spot captures transcripts from approximately 1–10 cells. After excluding spots not mapped to tissue regions, we retained 4,226, 4,384, 4,789, and 4,634 spots from tissue sections 151507–151510 (donor 1); 3,661, 3,498, 4,110, and 4,015 spots from sections 151669–151672 (donor 2); and 3,639, 3,673, 3,592, and 3,460 spots from sections 151673–151676 (donor 3).

\subsubsection*{STARmap mouse medial prefrontal cortex data} The mouse medial prefrontal cortex (mPFC) dataset generated using STARmap~\cite{wang2018three} was obtained from \url{https://figshare.com/articles/dataset/STARmap_datasets/22565200}. STARmap is an imaging-based ST technology that enables three-dimensional single-cell resolution profiling of gene expression within intact tissue. In this dataset, a targeted panel of 166 genes was measured across three mPFC sections, with 1,049 cells profiled in section BZ5, 1,053 cells in section BZ9, and 1,088 cells in section BZ14.

\clearpage

\bibliography{sample}

\clearpage

\section*{Author contributions}
R.Y. and L.X. conceived and designed the study; R.Y. conducted the experiment(s) and analyzed the results; R.Y prepared the first draft of the manuscript; R.Y., X.X., X.W, Z.Z, M.T.I and L.X. revised the mansucript; All authors contributed to manuscript preparation.

\section*{Acknowledgements}
The authors would like to thank the grant support from Stanford Human-Centered Intelligence (HAI) and the National Institutes of Health (NIH) (5R01CA256890, 1R01CA275772 (LX), and 1K99LM014309(TL)).

\section*{Competing Interests}
The authors declare that there are no competing interests.

\section*{Code availability}
Please visit \url{https://github.com/rui-yan/CellScape} for the source codes for code implementation, model training and evaluation.

\end{document}